\documentclass[runningheads]{eccv/llncs}
\PassOptionsToPackage{dvipsnames}{xcolor}
\usepackage{graphicx}
\usepackage{tikz}
\usepackage{eccv/comment}
\usepackage{amsmath,amssymb} % define this before the line numbering.
\usepackage{color}

\usepackage[accsupp]{axessibility}  % Improves PDF readability for those with disabilities.

\usepackage{multirow}
\usepackage{enumitem}
\usepackage[ruled]{algorithm2e}
\usepackage[noend]{algpseudocode}
\usepackage{tabularx}
\usepackage{booktabs}
\usepackage{caption}
\usepackage{subcaption}
\usepackage{makecell}
\usepackage{float}
\usepackage{hyperref}

\newcommand{\inlinesection}[1]{\noindent \textbf{#1}$\,\,\,$}

\def\methodname{MaskCLIP}

\def\ph{$\cdot$}

\newcommand{\newcontent}[1]{{\color{black}{#1}}} 

\usepackage[capitalize]{cleveref}
\crefname{section}{Sec.}{Secs.}
\Crefname{section}{Section}{Sections}
\Crefname{table}{Table}{Tables}
\crefname{table}{Tab.}{Tabs.}

\makeatletter
\def\blfootnote{\gdef\@thefnmark{}\@footnotetext}
\makeatother

\makeatletter
\DeclareRobustCommand\onedot{\futurelet\@let@token\@onedot}
\def\@onedot{\ifx\@let@token.\else.\null\fi\xspace}

\def\eg{\emph{e.g}\onedot} 
\def\ie{\emph{i.e}\onedot}

\def\etal{\emph{et al}\onedot}
\makeatother

\newif\ifsupp

\begin{document}
% \linenumbers
\pagestyle{headings}
\mainmatter
\def\ECCVSubNumber{910}  % Insert your submission number here

\title{Extract Free Dense Labels from CLIP} % Replace with your title

\begin{comment}
\titlerunning{ECCV-22 submission ID \ECCVSubNumber} 
\authorrunning{ECCV-22 submission ID \ECCVSubNumber} 
\author{Anonymous ECCV submission}
\institute{Paper ID \ECCVSubNumber}
\end{comment}
%******************

% \begin{comment}
\titlerunning{MaskCLIP}
% \author{Chong Zhou\inst{1}\orcidlink{0000-0002-9776-7739} \and
\author{Chong Zhou\inst{1} \and
Chen Change Loy\inst{1}\index{Loy, Chen Change} \and
Bo Dai\inst{2}\thanks{Bo Dai completed this work when he was with S-Lab, NTU.}}
\authorrunning{C. Zhou et al.}
\institute{S-Lab, Nanyang Technological University \and Shanghai AI Laboratory\\
\email{\{chong033, ccloy\}@ntu.edu.sg \quad daibo@pjlab.org.cn}}
% \end{comment}
%******************
\maketitle

\begin{abstract}
    Contrastive Language-Image Pre-training (CLIP) has made a remarkable breakthrough in open-vocabulary zero-shot image recognition. Many recent studies leverage the pre-trained CLIP models for image-level classification and manipulation.
    In this paper, we wish examine the intrinsic potential of CLIP for pixel-level dense prediction, specifically in semantic segmentation. 
    To this end, with minimal modification, we show that \methodname{} yields compelling segmentation results on open concepts across various datasets \textit{in the absence of annotations and fine-tuning}.
    By adding pseudo labeling and self-training, \methodname{}+ surpasses SOTA transductive zero-shot semantic segmentation methods by large margins, \eg, mIoUs of unseen classes on PASCAL VOC/PASCAL Context/COCO Stuff are improved from 35.6/20.7/30.3 to 86.1/66.7/54.7.
    We also test the robustness of \methodname{} under input corruption and evaluate its capability in discriminating fine-grained objects and novel concepts. Our finding suggests that \methodname{} can serve as a new reliable source of supervision for dense prediction tasks to achieve annotation-free segmentation. Source code is available \href{https://github.com/chongzhou96/MaskCLIP}{here}.
\end{abstract}
\section{Introduction}
\label{sec:intro}

% Through learning from massive data, 
Large-scale visual-language pre-training models such as CLIP~\cite{clip} capture expressive visual and language features. 
Various downstream vision tasks, \eg, text-driven image manipulation~\cite{style-clip}, image captioning~\cite{clip-score}, view synthesis~\cite{diet-nerf}, and object detection~\cite{ViLD}, have attempted to exploit such features for improved generality and robustness.
For instance, conducting zero-shot image classification based on raw CLIP features leads to a competitive approach that matches the performance of fully-supervised counterparts \cite{clip}.

\begin{figure}
\centering
\includegraphics[width=\textwidth]{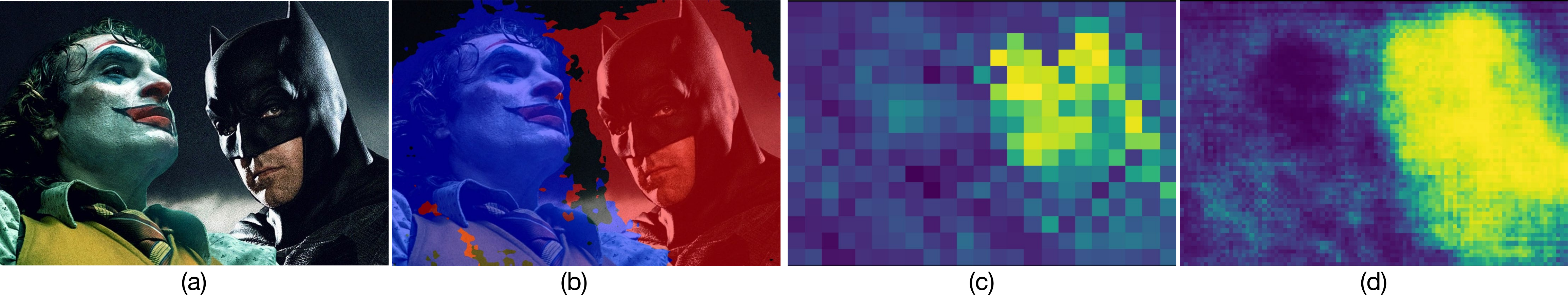}
\caption{Here we show the original image in (a), the segmentation result of \methodname+ in (b), and the confidence maps of \methodname~and \methodname+ for \emph{Batman} in (c) and (d) respectively. 
Through the adaptation of CLIP, \methodname~can be directly used for segmentation of fine-grained and novel concepts (e.g., \emph{Batman} and \emph{Joker}) without any training operations and annotations. Combined with pseudo labeling and self-training, \methodname+ further improves the segmentation result.}
\label{fig:teaser}
\end{figure}

In this paper, we take a step further to explore the applicability of CLIP features for pixel-level dense prediction tasks such as semantic segmentation.
This investigation is meaningful in that previous studies mainly leverage CLIP features as a global image representation. 
In contrast, our exploration wishes to ascertain the extent of CLIP features in encapsulating object-level and local semantics for dense prediction.
Different from the conventional pre-training task of image classification on iconic images, CLIP learns from images of complex scenes and their descriptions in natural language, which (1) encourages it to embed local image semantics in its features, (2) enables it to learn concepts in open vocabulary, and (3) captures rich contextual information, such as the co-occurrence/relation of certain objects and priors of the spatial locations. We believe all these merits contribute significantly to its potential for dense prediction tasks.

In this paper, we summarize both our success and failure experience on leveraging CLIP features for dense prediction. 
We find it essential to not break the visual-language association in the original CLIP feature space. In our earlier exploration, we experienced failures with the attempt to fine-tune the image encoder of CLIP for the segmentation task, \eg, initializing DeepLab \cite{deeplabv2} with the weights of CLIP's image encoder and fine-tune the backbone on segmentation. 
In addition, we found it is of utmost importance to avoid any unnecessary attempts to manipulate the text embeddings of CLIP. Such an approach would fail in segmenting unseen classes.

In our successful model, named \textbf{\methodname}, we show that one can simply extract dense patch-level features from the CLIP's image encoder, \ie, the \emph{value} features of the last attention layer, without breaking the visual-language association.
Classification weights for dense prediction, which are essentially 1$\times$1 convolutions, can be directly obtained from the text embeddings of CLIP's text encoder without any deliberate mapping.
In our empirical study, \methodname~yields reasonable predictions in both quantitative performance measured by mIoU metric and qualitative results.
Besides, \methodname~can be based on all variants of CLIP, including ResNets and ViTs. And we provide side-by-side comparisons between the two popular backbone networks. We also propose two mask refinement techniques for \methodname~to further improve its performance, namely \emph{key smoothing} and \emph{prompt denoising}, both require no training. Specifically, key smoothing computes the similarity between the \emph{key} features (of the last attention layer) of different patches, which are used to smooth the predictions. Prompt denoising removes prompts with classes that unlikely exist in the image, thus with fewer distractors, predictions become more accurate.

However, it is hard to further improve the segmentation capacity of \methodname{} as its architecture is restricted to be the image encoder of CLIP.
To relax \methodname{} from the architectural constraint and to incorporate more advanced architectures such as PSPNet \cite{pspnet} and DeepLab \cite{deeplabv2,deeplabv3+},
we notice that instead of deploying \methodname{} at the inference time,
we can deploy it at the training time,
where it serves as a generalizable and robust annotator 
that provides high-quality pseudo labels.
Together with a standard self-training strategy,
the resulting model,
termed \textbf{\methodname+}, achieves a strikingly remarkable performance.

Apart from annotation-free and open-vocabulary segmentation, \methodname+ can also be applied to the transductive zero-shot semantic segmentation task, where \methodname{} only generates pseudo labels for the unseen classes.
On the three standard segmentation benchmarks,
namely PASCAL VOC \cite{pascal-voc}, PASCAL Context \cite{pascal-context}, and COCO Stuff \cite{coco-stuff},
\methodname+ improves the state-of-the-art results in terms of mIoU of unseen classes, by 50.5\%, 46\%, and 24.4\%, respectively ($35.6\rightarrow86.1$, $20.7\rightarrow66.7$, and $30.3\rightarrow54.7$).
Thanks to the the generality and robustness of CLIP features,
\methodname+ can be readily applied to various extended settings of semantic segmentation,
including the segmentation of fine-grained classes (\eg,~attribute-conditioned classes like \emph{white car} and \emph{red bus}) or novel concepts (such as \emph{Batman} and \emph{Joker} as shown in Figure \ref{fig:teaser}), as well as the segmentation of moderately corrupted inputs. We show more interesting results in the experiment section.

Semantic segmentation is notorious for its high dependency on labeled training data. Many methods have been explored to get around such stringent requirement, \eg, through using weak labels like image tags, bounding boxes, and scribbles. Our study, for the first time, shows that features learned via large-scale visual-language pre-training can be readily used to facilitate open vocabulary dense prediction. The proposed model, MaskCLIP, shows promising potential in providing rich and meaningful dense pseudo labels for training existing methods.
\section{Related Work}
\inlinesection{Transferable Representation Learning.} Pre-training is widely used for dense prediction tasks. Yosinski \etal~\cite{transfer} show that ImageNet~\cite{imagenet} pre-training greatly speeds up the convergence of the downstream object detection task. Later, extensive research is conducted on making the pre-training a human-labor-free process. In particular, self-supervised representation learning constructs pretext tasks~\cite{pretask1,pretask2,pretask3} or relies on contrastive learning~\cite{moco,simclr,simsiam}, clustering\cite{swav}, and bootstrapping\cite{byol} to obtain supervision signals. Another line of work seeks to learn visual representation from natural language. Some studies \cite{berkeley-cap,self-cap,retrieval-cap,naver-cap,virtex,contrast-cap} propose to learn from image-caption pairs. Recently, CLIP~\cite{clip} and ALIGN~\cite{align} perform contrastive learning on very large-scale web-curated image-text pairs and show promising pre-trained representations with impressive zero-shot transferability. 
The success of CLIP inspires a new way of studies that transfer the pre-trained CLIP model to various downstream tasks such as text-driven image manipulation~\cite{style-clip}, image captioning~\cite{clip-score}, view synthesis~\cite{diet-nerf}, and object detection~\cite{ViLD}.
Different from these methods that typically apply CLIP right off the shelf for image encoding, we explore ways to adapt CLIP for pixel-level dense prediction.
A concurrent work, DenseCLIP~\cite{denseclip}, aims to better fine-tune the CLIP pre-trained weights on target semantic segmentation datasets without keeping the zero-shot transferability, which are different from our setting. To examine the intrinsic potential of CLIP for dense prediction tasks, we refrain from any fine-tuning and major architectural modification.

\inlinesection{Zero-Shot Visual Recognition.} Zero-shot learning aims at classifying instances of those categories that are not seen during training. Common clues to infer unseen categories include shared attributes and visual-semantic mapping. As the latter does not require extra annotations, the paradigm is well-suited for zero-shot dense prediction tasks. Zhao~\etal~\cite{open-parse} project image pixel features and word concepts into a joint space. Kato~\etal~\cite{iccv-workshop} fuse semantic features into visual features as guidance. ZS3Net~\cite{zs3} proposes to generate fake pixel-level features from semantic features for the unseen. SPNet~\cite{spnet} learns a projection from visual space to semantic space. Other studies like~\cite{uncertain-seg}, \cite{consistent-seg}, and \cite{context-seg}, improve the generative ZS3Net in terms of uncertainty, structural consistency, and context, respectively, while STRICT~\cite{cvpr-workshop} boosts the SPNet through self-training.
\newcontent{Depending on whether the unlabeled pixels are observed during training, the setting can be split into inductive (not observed) and transductive.}
We show that the proposed \methodname~not only achieves new SOTA on the zero-shot segmentation setting but can also deal with more difficult settings where all the categories are unseen during training.

\inlinesection{Self-Training.} Self-training leverages the model trained on labeled data to generate pseudo labels for the unlabeled, which then are used to iteratively improve the previous model. Self-training has firstly become popular in the semi-supervised classification task~\cite{pseudo-label,label-prop,learn-self,meta-pseudo} and is also recently applied in the semi-supervised/zero-shot semantic segmentation settings~\cite{adv-semi,consist-semi,naive-student,correct-semi,cross-consist-semi,error-semi,pseudo-seg,cross-semi}. Our \methodname+ adopts the same philosophy where the pseudo labels are obtained from both frozen \methodname~and \methodname+ itself.
\section{Methodology}

Our study serves as an early attempt that explores the applicability of CLIP features for pixel-level dense prediction tasks.
We start with a brief introduction of CLIP and a na\"{i}ves solution as the preliminary,
followed by presenting the proposed \methodname{} in detail.

\subsection{Preliminary on CLIP}

CLIP \cite{clip} is a visual-language pre-training method that learns both visual and language representations from large-scale raw web-curated image-text pairs. 
It consists of an image encoder $\mathcal{V}(\cdot)$ and a text encoder $\mathcal{T}(\cdot)$,
both jointly trained to respectively map the input image and text into a unified representation space.
CLIP adopts contrastive learning as its training objective,
where ground-truth image-text pairs are regarded as positive samples,
and mismatched image-text pairs are constructed as negative ones.
In practice, the text encoder is implemented as a Transformer \cite{transformer}.
As for the image encoder,
CLIP provides two alternative implementations,
namely a Transformer and a ResNet \cite{resnet} with global attention pooling layer.
Our method can be based on both encoder architectures. 

We believe CLIP has inherently embedded local image semantics in its features as it learns to associate image content with natural language descriptions, the latter of which contain complex and dense semantic guidance across multiple granularities.
For example,
to correctly identify the image corresponds to the description \emph{the man at bat readies to swing at the patch while the umpire looks on} \cite{coco-caption},
CLIP must divide image semantics into local segments and properly align image semantics with singular mentioned concepts like \emph{man, bat, swing, patch, man at bat, man at patch}, and \emph{man readies to swing},
instead of handling the image as a whole. Such uniqueness is absent from training with solely image labels.

\subsection{Conventional Fine-Tuning Hinders Zero-Shot Ability}
The current de facto pipeline of training a segmentation network is (1) initializing the backbone network with the ImageNet \cite{imagenet} pre-trained weights, (2) adding segmentation-specific network modules with randomly initialized weights, and (3) jointly fine-tuning the backbone and newly added modules.

It is natural to follow these standard steps to adapt CLIP for segmentation. Here, we start our exploration by applying this pipeline on DeepLab \cite{deeplabv2} with two CLIP-specific modifications.
Specifically,
we first replace the ImageNet pre-trained weights with weights of the image encoder of CLIP. 
Second, we adopt a mapper $\mathcal{M}$ that maps text embeddings of CLIP to the weights of DeepLab classifier (the last $1\times1$ convolutional layer).
The modified model can be formulated as follows:
\begin{align}
\text{DeepLab}(x) &=\mathcal{C}_{\phi}(\mathcal{H}(\mathcal{V}_{*l}(x))), \\
\phi &= \mathcal{M}(t),
\end{align}
where $\mathcal{V}_{*l}(\cdot)$ denotes the DeepLab backbone, which is a ResNet dilated by a factor of $l$.
$H(\cdot)$ denotes the randomly initialized ASPP module~\cite{deeplabv2},
and $\mathcal{C}_{\phi}(\cdot)$ is the DeepLab classifier,
whose weights, denoted as $\phi$, are determined by the text embedding of CLIP via the mapper $\mathcal{M}$.
Ideally, by updating the classifier weights with the corresponding text embedding, the adapted DeepLab is able to segment different classes without re-training.

To evaluate the segmentation performance of this modified DeepLab on both seen and unseen classes,
we train it on a subset of classes in the dataset,
considering the remaining classes as unseen ones.
We have tried a series of mapper architectures. Although they perform well on seen classes,
in all these cases the modified DeepLab fails to segment unseen classes with satisfying performance.
We hypothesize that this is mainly because the original visual-language association of CLIP features has been broken: (1) the backbone is slightly different from the image encoder in terms of network architecture;
(2) weights initialized from the image encoder have been updated during fine-tuning;
(3) an extra mapper, 
which is trained only on data of seen classes, is introduced therefore leading to insufficient generality.

\begin{figure*}[t]
\centering
\includegraphics[width=\textwidth]{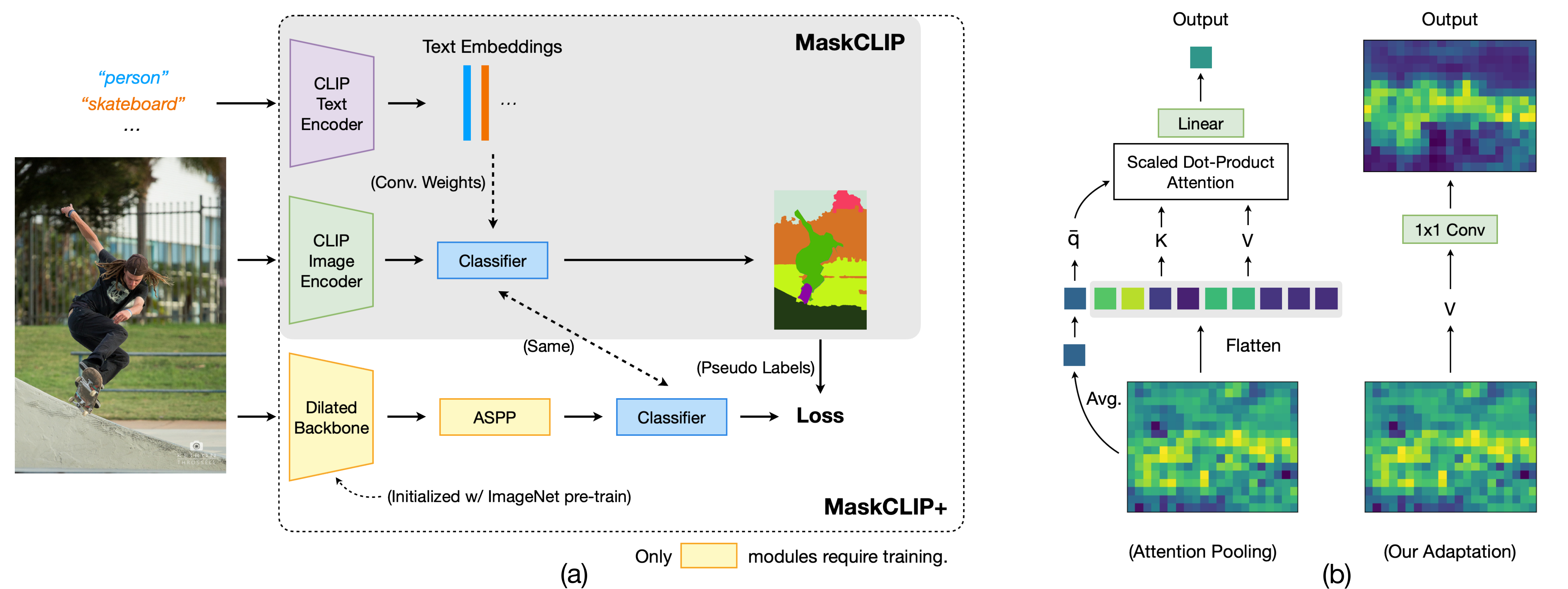}
\caption{\inlinesection{Overview of \methodname/\methodname+.} Compared to the conventional fine-tuning method, the key to the success of \methodname~is keeping the pre-trained weights frozen and making minimal adaptation to preserve the visual-language association. Besides, to compensate for the weakness of using the CLIP image encoder for segmentation, which is designed for classification, \methodname+ uses the outputs of \methodname~as pseudo labels and trains a more advanced segmentation network such as DeepLabv2~\cite{deeplabv2}}
\label{fig:architecture}
\end{figure*}

\subsection{\methodname}
Failing the fine-tuning attempt, we turn to a solution that avoids introducing additional parameters and modifying the feature space of CLIP.
To this end,
we carefully revisit the image encoder of CLIP, especially its unique global attention pooling layer. 
As shown in Figure \ref{fig:architecture}(b), 
different from conventional global averaged pooling,
the image encoder of CLIP adopts a Transformer-style multi-head attention layer 
where globally average-pooled feature works as the query, and feature at each spatial location generates a key-value pair. 
Consequently,
the output of this layer is a spatial weighted-sum of the incoming feature map followed by a linear layer $\mathcal{F}(\cdot)$:
\begin{align}
\text{AttnPool}(\bar{q}, k, v) &= \mathcal{F}(\sum_{i}\text{softmax}(\frac{\bar{q}k_{i}^{\mathsf{T}}}{C})v_{i}) \notag \\
                               &= \sum_{i}\text{softmax}(\frac{\bar{q}k_{i}^{\mathsf{T}}}{C})\mathcal{F}(v_{i}),\label{eq:attn-pool} \\
\bar{q}=\text{Emb}_\text{q}(\bar{x}),\, k_i&=\text{Emb}_\text{k}(x_i),\, v_i=\text{Emb}_\text{v}(x_i),
\end{align}
where $C$ is a constant scaling factor and Emb($\cdot$) denotes a linear embedding layer\footnote{Here we have simplified the formula by ignoring the channel-wise splitting and concatenation.}. $x_i$ represents the input feature at spatial location $i$ and $\bar{x}$ is the average of all $x_{i}$.
The outputs of the Transformer layer serve as a comprehensive representation of the whole image. We believe that this is possible because $\mathcal{F}(v_i)$ computed at each spatial location already captures a rich response of local semantics that correspond well with tokens in the text embeddings of CLIP.

Based on such a hypothesis, 
as shown in Figure \ref{fig:architecture}(b), we directly modify the image encoder of CLIP in our new attempt:
(1) removing the query and key embedding layers;
(2) reformulating the value-embedding layer and the last linear layer into two respective $1\times1$ convolutional layers. 
Moreover, we keep the text encoder unchanged and it takes prompts with target classes as the input. The resulting text embedding of each class is used as the classifier.
We name the resulting model as \methodname~since it yields pixel-level mask predictions instead of a global image-level prediction.
We then evaluate \methodname~on various standard segmentation benchmarks as well as web-crawled images. 
As shown in Figure \ref{fig:teaser}, \methodname~can output reasonable results without any fine-tuning nor annotations. 
More qualitative results and quantitative results with respect to the mIoU metric are included in the experiment section.

One might argue that, since the global attention pooling is a self-attention layer, even without modification, it can also generate dense features. However, since query $\bar{q}$ is the only query trained during the CLIP pre-training, this na\"{i}ves solution fails. We treat this solution as the baseline and compare its results with ours in the experiments.
Moreover, the Transformer layer in ViT is very similar to the global attention pooling. In fact, the only two differences are: (1) the global query is generated by a special [CLS] token instead of the average among all spatial locations; (2) Transformer layer has a residual connection. Therefore, by replacing $\bar{q}$ with $q_{[cls]}$ and adding input $x$ to the output, \methodname{} can work with the ViT backbone.

Despite the simplicity of \methodname~in comparison to existing segmentation approaches, the proposed method enjoys multiple unique merits inherited from CLIP. 
First, \methodname~can be used as a free segmentation annotator to provide rich and novel supervision signals for segmentation methods working with limited labels. 
Second, since the visual-language association of CLIP is retained in \methodname, 
it naturally possesses the ability to segment open vocabulary classes, as well as fine-grained classes described by free-form phrases,
such as \emph{white car} and \emph{red bus}.
Third, since the CLIP is trained on raw web-curated images, CLIP demonstrates great robustness to natural distribution shift~\cite{clip} and input corruptions \cite{inverse}.
We verify that \methodname~preserves such robustness to some extent.

\inlinesection{Key Smoothing and Prompt Denoising.} To further improve the performance of \methodname, we propose two refinement strategies, namely \emph{key smoothing} and \emph{prompt denoising}.
Recall that, in Eq.~\ref{eq:attn-pool}, besides $\bar{q}$, key features $k_i$ also get trained during CLIP pre-training. However, in the original \methodname, $k_i$ is simply discard. Hence, here we seek to utilize this information to refine the final output.
Key features can be viewed as the descriptor of the corresponding patch, therefore patches with similar key features should yield similar predictions. With this hypothesis, we propose to smooth the predictions with:
\begin{align}
\text{pred}_i &= \sum_{j}\text{cos}(\frac{k_i}{\lVert k_i \rVert_2}, \frac{k_j}{\lVert k_j \rVert_2})\text{pred}_i,
\end{align}
where $k_i$ and $\text{pred}_i$ are key features and class confidence predictions at spatial location $i$, while $\lVert \cdot \rVert_2$ and $\text{cos}(\cdot)$ denote L2 normalization and cosine similarity. We name this strategy as key smoothing.

In addition, we also observe that when dealing with many target classes, since only a small proportion of the classes appear in a single image, the rest classes are in fact distractors and undermine the performance. Therefore, we propose prompt denoising, which removes the prompt with target class if its class confidence at all spatial locations is all less than a threshold $t=0.5$.

\subsection{\methodname+}
While \methodname~does not require any training, its network architecture is rigid because it adopts the image encoder of CLIP.
To relax it from this constraint and benefit from more advanced architectures tailored for segmentation, such as DeepLab \cite{deeplabv2} and PSPNet \cite{pspnet}, 
we propose \methodname+.
Instead of directly applying \methodname~for test-time prediction,
\methodname+~regard its predictions as training-time pseudo ground-truth labels.
Together with an adopted self-training strategy,
\methodname+~is thus free from the restriction on its backbone architecture.
As shown in Figure \ref{fig:architecture}(a),
we take DeepLabv2~\cite{deeplabv2} as the backbone of \methodname+~to ensure a fair comparison with previous segmentation methods.

\inlinesection{\methodname-Guided Learning.}
In \methodname+, we leverage the predictions of \methodname~to guide the training of another target network comprising an architecture tailored to segmentation task.
In parallel to the target network, we feed the same pre-processed image input to the \methodname~and use the predictions of \methodname~as pseudo ground-truth labels to train the target network.
In addition, we replace the classifier of the target network with that of \methodname, to preserve the network's ability for open vocabulary prediction.

\methodname-guided learning is also applicable in the transductive zero-shot segmentation setting. 
Specifically, while pixels of both seen and unseen classes are observed, only annotations of seen classes are available.
In this case, we only use \methodname~to generate pseudo labels for the unlabeled pixels. 
Compared to SOTA methods, \methodname+ obtains remarkably better results across three standard benchmarks, namely PASCAL VOC 2012 \cite{pascal-voc}, PASCAL Context \cite{pascal-context}, and COCO Stuff \cite{coco-stuff},
where the results of \methodname+ are even on par with that of fully-supervised baselines.

We note that some related attempts \cite{ViLD,zsd-yolo}, targeting object detection, perform knowledge distillation between the image-level visual features of CLIP and the features of a target model.
Different from such feature-level guidance,
we adopt pseudo labels in our case.
This is because our target network, with a segmentation-tailored architecture, is structurally distinct from the image encoder of CLIP.
Therefore, distillation by feature matching may be a sub-optimal strategy.
In fact, as reported by \cite{ViLD},
under zero-shot setting, such feature-level guidance indeed results in conflicts between the performance of seen and unseen classes.
On the contrary, 
by adopting pseudo labels in \methodname+,
we do not observe any performance drop on seen classes.

\inlinesection{Self-Training.} 
It is expected that after certain training iterations, the target network guided by \methodname~will outperform \methodname, 
rendering the latter suboptimal for further guidance as it gradually becomes an inferior model over time.
Empirically, we also find that \methodname-guided learning reaches an upper bound at around 1/10 of the standard training schedule. 
To further improve the performance, we swap out \methodname~and let the target model generate pseudo labels for itself. 
This is commonly referred to as self-training.
\section{Experiments}

\inlinesection{Datasets.}
We conduct experiments on three standard segmentation benchmarks, namely PASCAL VOC 2012 \cite{pascal-voc}, PASCAL Context \cite{pascal-context}, and COCO Stuff \cite{coco-stuff}. 
PASCAL VOC 2012 contains 1,426 training images with 20 object classes plus a background class. Following common practice, we augment it with the Semantic Boundaries Dataset~\cite{sbd}. PASCAL Context labels PASCAL VOC 2010 (4,998/5,105 train/validation images) with segmentation annotations of 520 object/stuff classes, from which the most common 59 classes are treated as foreground while the rest are regarded as background. COCO Stuff extends the COCO dataset, which contains segmentation annotations of 80 object classes on 164K images, with additional 91 stuff classes.

\inlinesection{Text Embedding.}
We follow the same process to construct text embeddings as Gu~\etal~\cite{ViLD}. Specifically, we feed prompt engineered texts into the text encoder of CLIP with 85 prompt templates, such as \emph{there is a \{class name\} in the scene},
and average the resulting 85 text embeddings of the same class.

\begin{table}[t]
\centering
{
\caption{\inlinesection{Annotation-free segmentation (mIoU).} \textbf{(a)} We evaluate the performance of \methodname(+) on two standard datasets. For Pascal Context, we ignore the evaluation on the background class. The target model of \methodname+ is Deeplabv2-ResNet101. KS and PD denote key smoothing and prompt denoising respectively. And they are not used in \methodname+. \textbf{(b)} We test the robustness of \methodname{} on Pascal Context under various types of corruption}
\begin{subfigure}[t]{0.5\textwidth}
    \caption{}\label{tab:no-anno}
    \begin{tabular}{l l c c }
        \toprule
        Method                          & CLIP      & \makecell{PASCAL\\Context} & \makecell{COCO\\Stuff} \\
        \midrule
        \multirow{2}{*}{Baseline}       & r50       &  8.3      &  4.6 \\
                                        & vit16     &  9.0      &  4.3 \\
        \midrule
        \multirow{6}{*}{\methodname}    & r50       & 18.5      & 10.2 \\
                                        & +KS       & 21.0      & 12.4 \\
                                        & +PD       & 19.0      & 10.8 \\
                                        & +KS+PD    & 21.8      & 12.8 \\
                                        \noalign{\medskip}
                                        & vit16     & 21.7      & 12.5 \\
                                        & +KS       & 23.9      & 13.8 \\
                                        & +PD       & 23.1      & 13.2 \\
                                        & +KS+PD    & 25.5      & 14.6 \\
        \midrule
        \multirow{2}{*}{\methodname+}   & r50       & 23.9     & 13.6 \\
                                        & vit16     & 31.1     & 18.0 \\
        \bottomrule
    \end{tabular}
\end{subfigure}
}
\,\,\,
\begin{subfigure}[t]{0.4\textwidth}
    \caption{} 
    \label{tab:bad-input}
    \begin{tabular}{l c cc c cc}
        \toprule
        \multirow{2}{*}{Corruption} & \multicolumn{2}{c}{\textbf{level 1}}  & \multicolumn{2}{c}{\textbf{level 5}}  \\
                                    & r50   & vit16                  & r50 & vit16 \\
        \midrule
        None                        & 18.5   & 21.7                  & 18.5   & 21.7 \\
        \midrule
        Gaussian Noise              & 13.7   & 19.6                  &  2.1   &  6.8 \\
        Shot Noise                  & 14.0   & 19.6                  &  2.4   &  7.5 \\
        Impulse Noise               &  9.9   & 17.3                  &  2.1   &  7.2 \\
        Speckle Noise               & 15.1   & 20.0                  &  5.6   & 11.4 \\
        \midrule
        Gaussian Blur               & 17.4   & 21.6                  &  4.3   & 14.1 \\
        Defocus Blur                & 15.7   & 20.8                  &  6.6   & 15.5 \\
        \midrule
        Spatter                     & 17.1   & 20.5                  &  7.8   & 12.2 \\
        JPEG                        & 15.7   & 20.8                  &  7.6   & 14.5 \\
        \bottomrule
    \end{tabular}
\end{subfigure}
\end{table}

\inlinesection{Implementation Details.} We implement our method on the \emph{MMSegmentation}\footnote{\url{https://github.com/open-mmlab/mmsegmentation}} codebase and inherit its training configurations. Input resolutions are set as 512x512. When using ViT, the pre-trained positional embeddings adopt bicubic interpolation. \methodname~requires no training and we train \methodname+ on 4 Tesla V100 GPUs with a batch size of 16. For annotation-free segmentation, we perform \methodname-guided learning for 4k/8k iterations on PASCAL Context/COCO Stuff with DeepLabv2-ResNet101 as the backbone segmentor. Self-training is not used in this setting. For zero-shot segmentation, we choose the lightest training schedule provided by \emph{MMSegmentation}, which is 20k/40k/80k for PASCAL VOC/PASCAL Context/COCO Stuff. The first 1/10 training iterations adopt \methodname-guided learning and the rest adopts self-training. For fair comparisons, we choose DeepLabv2 as the target model for PASCAL VOC and COCO Stuff and DeepLabv3+ for PASCAL Context. All use the ResNet-101 backbone initialized with the ImageNet pre-trained weights. 
Finally, we use the publicly available CLIP-ResNet-50 and CLIP-ViT-B/16 models\footnote{\url{https://github.com/openai/CLIP}}.

\subsection{Annotation-Free Segmentation}
In this challenging setting, no annotation is provided during training. We first evaluate the mIoU performance on two standard datasets, PASCAL Context and COCO-Stuff. Then we collect images from Flickr to show interesting qualitative results on novel concepts, such as \emph{Batman} and \emph{Joker}. Finally, we test the robustness of \methodname{} under various image corruptions.

\begin{figure}[t]
\centering
\includegraphics[width=\textwidth]{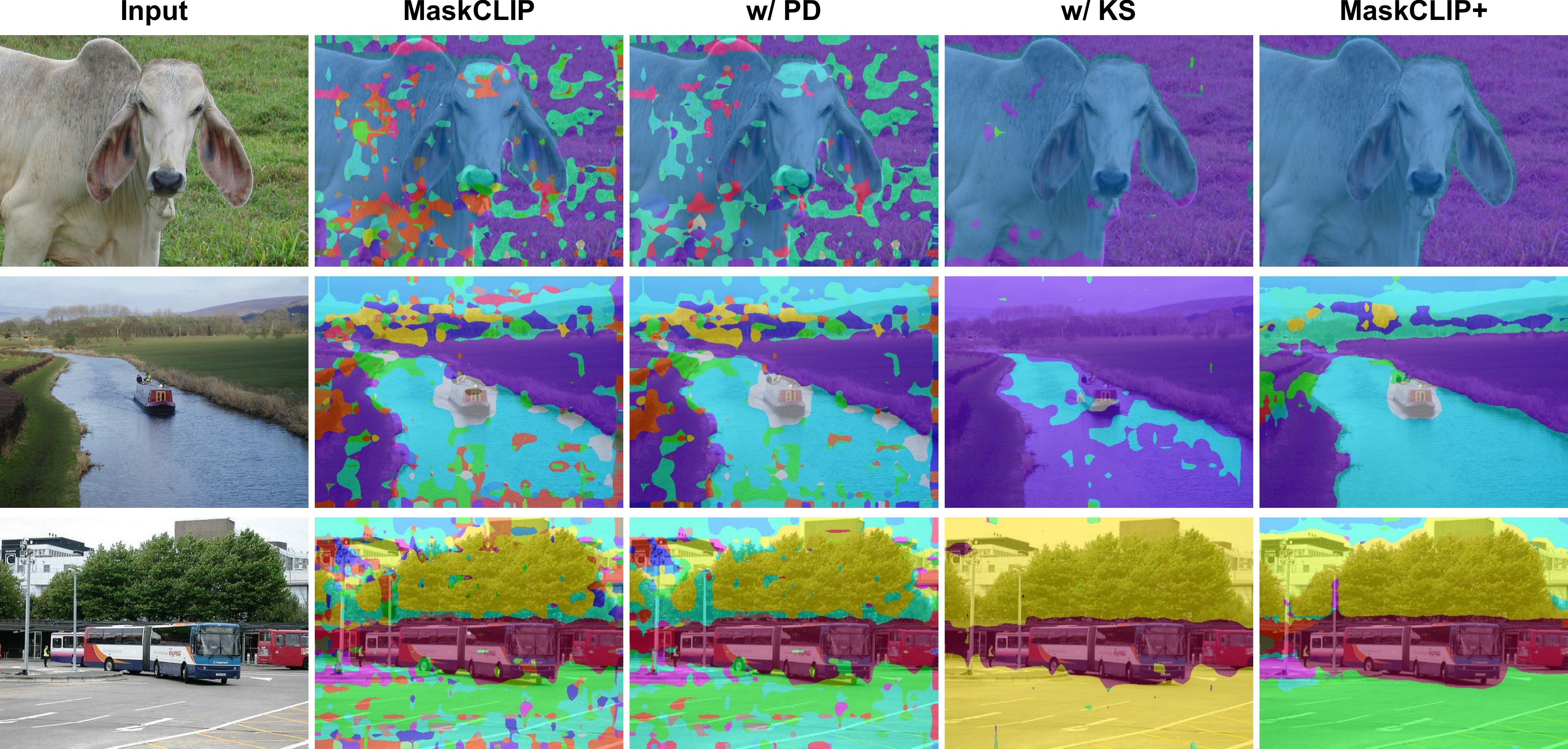}
\caption{\inlinesection{Qualitative results on PASCAL Context.} Here all results are obtained \textbf{without} any annotation. PD and KS refer to prompt denoising and key smoothing respectively. With PD, we can see some distraction classes are removed. KS is more aggressive. Its outputs are much less noisy but are dominated by a small number of classes. Finally, \methodname+ yields the best results} 
\label{fig:quality-ablation}
\end{figure}

\inlinesection{Performance on Standard Datasets.} In Table \ref{tab:no-anno}, we show mean Intersection over Union (mIoU) results on PASCAL Context and COCO-Stuff. The baseline in the table refers to directly using dense features from the CLIP's image encoder without any modification. As shown in the table, \methodname{} outperforms the baseline by huge margins,
indicating it is essential to avoid computing attention of the last attention layer and instead value features should be directly used.
The results also show that key smoothing and prompt denoising are effective and are orthogonal to each other.
Therefore, we empirically conclude that for each spatial location, the query features should be discarded and key/value features can be re-organized into final predictions. Furthermore, with the predictions of \methodname{} as pseudo labels, \methodname+ significantly improves the performance, \eg, on PASCAL Context, without any human annotation, \methodname+(ViT-B/16) obtains 31.1 mIoU. One may notice that ViT almost consistently surpasses ResNet. Apart from ViT-B/16 has more FLOPs than ResNet-50, another possible reason is that ViT only downsamples the input by 16 times whereas ResNet downsamples 32 times, which particularly matters for dense prediction tasks. Besides quantitative results, in Figure \ref{fig:quality-ablation}, we also visualize the outputs of each \methodname{} variant.

\begin{figure}[t]
\centering
\includegraphics[width=\textwidth]{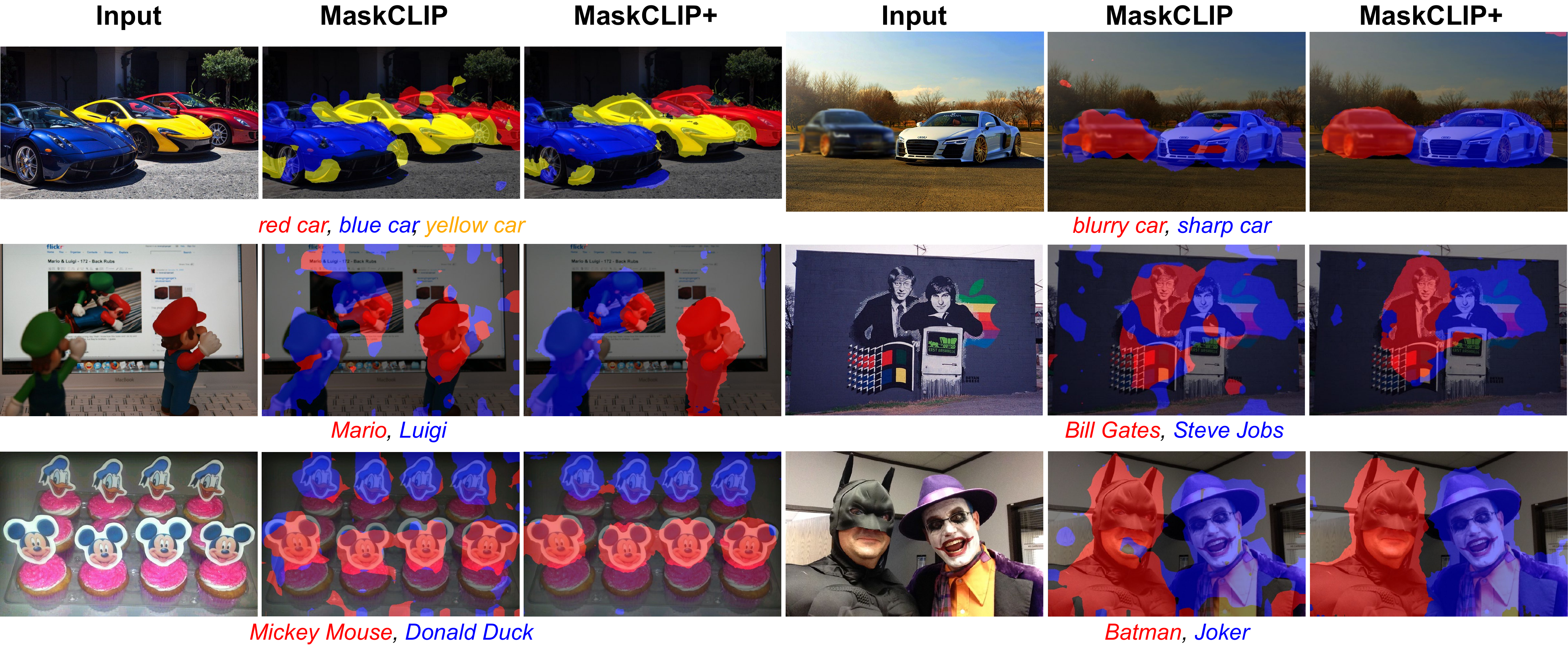}
\caption{\inlinesection{Qualitative results on Web images.} Here we show the segmentation results of \methodname~and \methodname+ on various \textbf{unseen classes}, including fine-grained classes such as cars in different colors/imagery properties, celebrities, and animation characters. All results are obtained \textbf{without} any annotation} 
\label{fig:quality}
\end{figure}

\inlinesection{Open-Vocabulary Segmentation on Web-Crawled Images.} \methodname~inherits the open-vocabulary ability from CLIP and does not require annotations. Therefore, we can deploy it on several interesting setups where the target classes are (1) more fine-grained, such as \emph{red car, yellow car}; (2) of certain imagery properties, \eg, blurry; (3) novel concepts like \emph{Batman, Joker}. To this end, we collect images from Flickr then directly evaluate these images on \methodname{} and train \methodname+ with only \methodname-guided learning. Note that, for the background, we enumerate a set of classes that might appear in the background and regard them as a whole as the background class. Results in Figure \ref{fig:quality} are impressive given the open-vocabulary targets and being annotation-free. Besides, results from \methodname+ are less noisy and more accurate than \methodname, which is complementary to the quantitative results.

\inlinesection{Robustness Under Corruption.} CLIP is trained on web-curated images, whose quality and distribution are more diverse than well-pre-processed datasets. Radford~\etal~\cite{clip} and Ravula~\etal~\cite{inverse} demonstrate the robustness of CLIP on natural distribution shift and artificial corruption respectively. While these explorations are done for image classification, we benchmark its robustness for dense prediction tasks. Specifically, we impose various corruptions used in ImageNet-C \cite{imagenet-c} with different severity levels on images in PASCAL Context and evaluate on \methodname. In Table \ref{tab:bad-input}, \methodname~models based on CLIP-ViT-B/16 are much more robust than CLIP-ResNet-50. In particular, CLIP-ViT-B/16 rarely suffers from degradation across a wide range of corruptions with level 1 severity and is cable of generating reasonable labels even under the most severe corruptions (level 5\footnote{The severity level is controlled by certain coefficients, such as kernel size, specified in ImageNet-C \cite{imagenet-c}.}).

\begin{table}[t]
\centering

{
\def\sp{SPNet}
\def\spc{SPNet-C}
\def\zs{ZS3Net}
\def\cag{CaGNet}
\def\spst{SPNet+ST}
\def\zsst{ZS3Net+ST}
\def\cagst{CaGNet+ST}
\def\strict{STRICT}
\def\denseclip{\methodname}
\def\denseclipplus{\methodname+}
\def\voc{\textbf{PASCAL-VOC}}
\def\context{\textbf{PASCAL-Context}}
\def\stuff{\textbf{COCO-Stuff}}
\newcommand{\better}[1]{\textcolor{ForestGreen}{+#1}}

\caption{\inlinesection{Zero-shot segmentation performances.} ST stands for self-training. mIoU(U) denotes mIoU of the unseen classes. \spc~is the SPNet with calibration. 
% The background class in PASCAL VOC is ignore, while that in PASCAL Context is not ignored. 
On PASCAL Context, all methods use DeepLabv3+-ResNet101 as the backbone segmentation model and the rest two datasets use DeepLabv2-ResNet101. For \methodname+, CLIP-ResNet-50 is used to generate pseudo labels}
\label{tab:zero-shot}
\setlength{\tabcolsep}{1pt}
\begin{tabular}{l ccc c ccc c ccc}
    \noalign{\medskip}
    \toprule
    \multirow{2}{*}{Method} & \multicolumn{3}{c}{\voc}     && \multicolumn{3}{c}{\stuff}       && \multicolumn{3}{c}{\context} \\
                            &  mIoU(U) & mIoU & hIoU       && mIoU(U) & mIoU & hIoU            && mIoU(U) & mIoU & hIoU \\
    \midrule
    Inductive \\
    \midrule
    \sp                     & 0.0 & 56.9 & 0.0                && 0.7 & 31.6 & 1.4          && \ph & \ph & \ph \\
    \spc                    & 15.6 & 63.2 & 26.1            && 8.7 & 32.8 & 14.0         && \ph & \ph & \ph \\
    \zs                     & 17.7 & 61.6 & 28.7            && 9.5 & 33.3 & 15.0         && 12.7 & 19.4 & 15.8 \\
    \cag                    & 26.6 & 65.5 & 39.7            && 12.2 & 33.5 & 18.2        && 18.5 & 23.2 & 21.2 \\
    \midrule
    Transductive \\
    \midrule
    \spst                   & 25.8 & 64.8 & 38.8            && 26.9 & 34.0 & 30.3        && \ph & \ph & \ph \\
    \zsst                   & 21.2 & 63.0 & 33.3            && 10.6 & 33.7 & 16.2        && 20.7 & 26.0 & 23.4 \\
    \cagst                  & 30.3 & 65.8 & 43.7            && 13.4 & 33.7 & 19.5        && \ph & \ph & \ph \\
    \strict                 & 35.6 & 70.9 & 49.8            && 30.3 & 34.9 & 32.6        && \ph & \ph & \ph \\
    \denseclipplus          & \textbf{86.1} & \textbf{88.1} & \textbf{87.4}
                            && \textbf{54.7} & \textbf{39.6} & \textbf{45.0} 
                            && \textbf{66.7} & \textbf{48.1} & \textbf{53.3} \\
                            & \better{50.5} & \better{17.2} & \better{37.6}    
                            && \better{24.4} & \better{4.7}  & \better{12.4}        
                            && \better{46.0} & \better{22.1} & \better{29.9} \\
    \midrule
    Fully Sup.              & \ph & 88.2 & \ph              && \ph & 39.9 & \ph           && \ph & 48.2 & \ph \\
    \bottomrule
\end{tabular}

}
\end{table}

\subsection{Zero-Shot Segmentation}
Apart from annotation-free segmentation, \methodname+ can also be applied to the zero-shot segmentation task with minor effort. Specifically, in the zero-shot setting, pixels of certain classes do not have annotations, to which \methodname{} can assign reliable pseudo labels.

\inlinesection{Zero-shot Setups.} Traditionally, zero-shot segmentation methods train on a subset of classes, named seen classes, with ground-truth annotations, and during inference, both seen and unseen classes are evaluated. Depending on whether the unlabeled pixels are observed during training, the setting can be split into inductive (not observed) and transductive (observed). Our method conforms to the transductive setting.

The selection of seen classes varies among previous works and we follow the most common setups, where for PASCAL VOC, the background class is ignored and \emph{potted plant, sheep, sofa, train, tv monitor} are chosen as the 5 unseen classes; for PASCAL Context, the background is not ignored and \emph{cow, motorbike, sofa, cat, boat, fence, bird, tv monitor, keyboard, aeroplane} are unseen; and for COCO Stuff, \emph{frisbee, skateboard, cardboard, carrot, scissors, suitcase, giraffe, cow, road, wall concrete, tree, grass, river, clouds, playing field} are unseen. 
We report the mean Intersection over Union (mIoU) of seen, unseen, and all classes as well as the harmonic mean (hIoU) of seen and unseen mIoUs as evaluation metrics.

We compare \methodname+ with SOTA methods including SPNet~\cite{spnet}, ZS3Net \cite{zs3}, CaGNet~\cite{context-seg}, and STRICT~\cite{cvpr-workshop}.
ZS3Net and CaGNet are generative approaches, while SPNet is non-generative and more simple but requires post-possessing step of calibration (SPNet-C). STRICT improves SPNet by a self-training strategy and is free of calibration. Compare with these methods, our \methodname+ does not rely on any particular network architecture nor post-possessing. Note that similar to CLIP, all methods, except for ZS3Net, do not exclude unseen classes during pre-training. Besides, \methodname+ also follows the rule that pixel-level annotations of unseen classes are prohibited. Thus, the comparison is fair.

Despite being simple, \methodname+ achieves a strikingly good result. As shown in Table \ref{tab:zero-shot}, it surpasses all methods on all datasets with large margins. On PASCAL VOC, PASCAL Context, and COCO Stuff, in terms of unseen mIoUs, \methodname+ improves the previous SOTA by 50.5, 24.4, and 46.0 respectively (on a scale of 100).
Note that the overall mIoU of \methodname+ is on par with that of fully supervised baselines.
Please refer to Table \ref{tab:zero-shot} for more specific numbers.

\begin{table}[t]
    \centering
    \caption{\inlinesection{Ablations of \methodname+.} Experiments are performed on the PASCAL VOC dataset under the zero-shot setting 
    % (the background class is ignored)
    }
    \label{tab:ablations}
    \begin{tabular}{l c c c c}
        \noalign{\medskip}
        \toprule
        Method                  & mIoU(S)           & mIoU(U)       & mIoU              & hIoU \\
        \midrule
        Adapted DeepLabv2       & 83.4              & 3.7           & 63.5              & 7.0 \\
        + \methodname-Guided    & \textbf{89.5}     & 72.8          & 85.3              & 80.3 \\
        + Self-Training         & 88.8              & \textbf{86.1} & \textbf{88.1}     & \textbf{87.4} \\
        \bottomrule
    \end{tabular}
\end{table}

\inlinesection{Ablation Studies of \methodname+.} We perform ablation studies on the PASCAL VOC zero-shot segmentation setting. As shown in Table \ref{tab:ablations}, we first examine the two proposed strategies in \methodname+. Compared to the adapted DeepLabv2, whose classifier is replaced with the \methodname{} classifier, \methodname-guided learning improves the unseen mIoU from 3.7 to 72.8 and the result is further improved by self-training to 86.1. However, there is a slight degradation on seen classes when using self-training (from 89.5 to 88.8) partially due to model drifting. Overall, \methodname+ performs better than \methodname{} on unseen classes and surpasses the baseline DeepLabv2 on seen classes in the same time.
\section{Conclusion}
% In this paper,
This paper presents our exploration of applying CLIP in semantic segmentation,
as an early attempt that studies the applicability of pre-trained visual-language models in pixel-level dense prediction tasks. 
While the conventional fine-tuning paradigm fails to benefit from CLIP, 
we find the image encoder of CLIP already possesses the ability to directly work as a segmentation model.
The resulting model, termed \methodname, can be readily deployed on various semantic segmentation settings without re-training. 
% Two useful training-free approaches, namely key smoothing and prompt denoising, are presented to further improve the performance.
On top of the success of \methodname,
we further propose \methodname+ that leverages \methodname~to provide training-time pseudo labels for unlabeled pixels,
which thus can be applied to more segmentation-tailored architectures beyond just the image encoder of CLIP.
On standard transductive zero-shot segmentation benchmarks,
\methodname+ significantly improves previous SOTA results.
% even doubles the mIoU of unseen classes.
More importantly,
\methodname+ can be readily employed for segmenting more challenging unseen classes, 
% including fine-grained concepts.
such as celebrities and animation characters.
%We verify the effectiveness of \methodname+ on transductive zero-shot segmentation and achieve new SotA across PASCAL VOC/PASCAL Context/COCO Stuff by large margins. We also conduct robustness tests and apply \methodname~on interesting settings, such as segmenting fine-grained classes and novel concepts. Our finding shows promising potentials of \methodname~used as a new reliable supervision source for dense prediction tasks.
\\~\\
\noindent\textbf{Acknowledgement.} This study is supported under the RIE2020 Industry Alignment Fund – Industry Collaboration Projects (IAF-ICP) Funding Initiative, as well as cash and in-kind contribution from the industry partner(s). This study is also supported by Singapore MOE AcRF Tier 2 (MOE-T2EP20120-0001) and Shanghai AI Laboratory.
\appendix
%%%%%%%% QUALITATIVE RESULTS %%%%%%%%
\ifsupp
\begin{figure}[H]
    \centering
    \includegraphics[width=\textwidth]{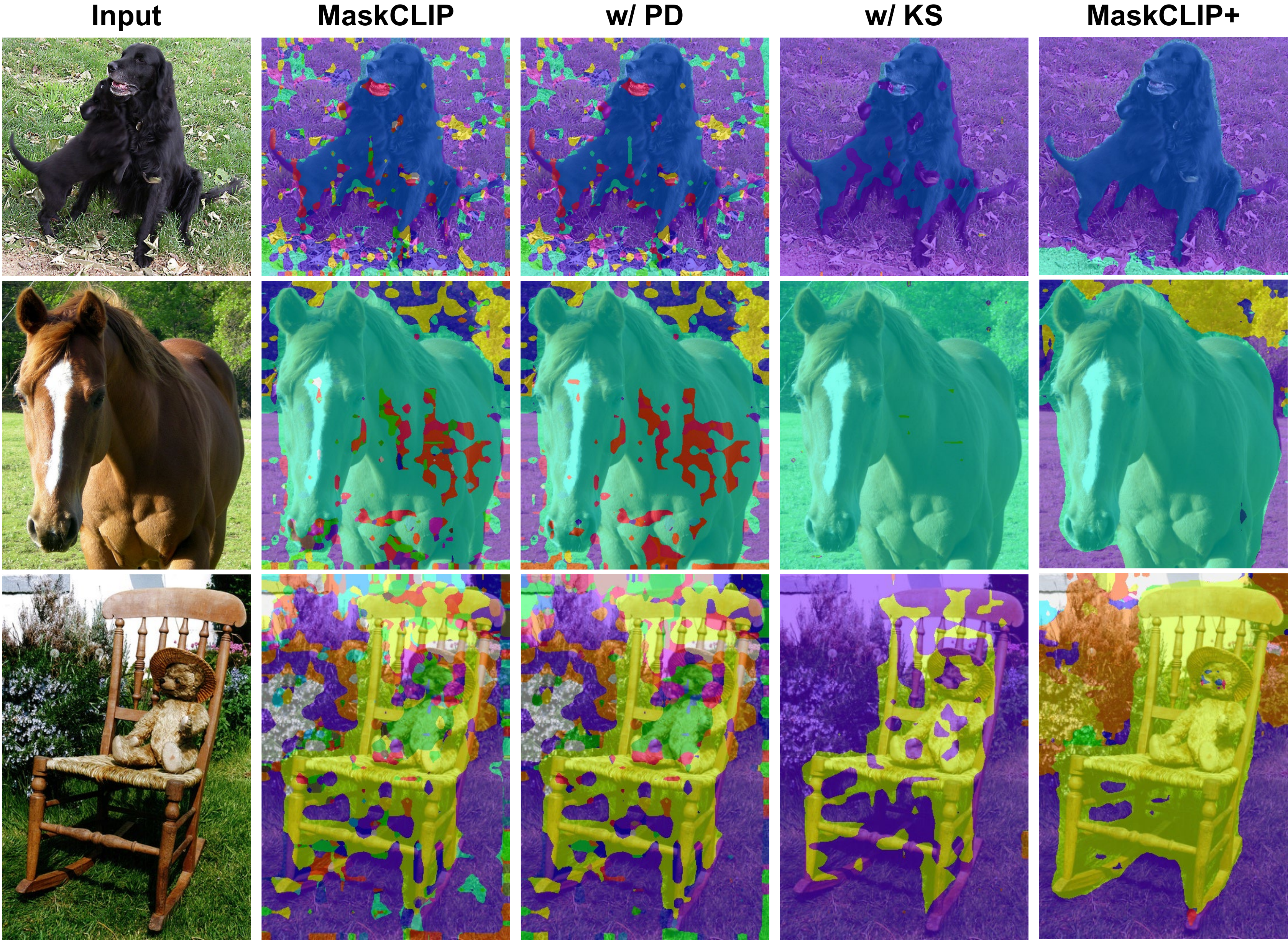}
    \caption{\inlinesection{More qualitative results on PASCAL Context.} Here all results are obtained \textbf{without} any annotation. PD and KS refer to prompt denoising and key smoothing respectively. Row 2, Col 4 shows a failure case of KS, where all the pixels in the image are labeled as the \emph{horse}. Note that, PASCAL Context does not contain \emph{bear} or \emph{teddy bear} classes and \methodname{} predicts the \emph{teddy bear} pixels as \emph{bedclothes}}
    \label{fig:supp-quality-abl}
    \vspace{-7pt}
\end{figure}

\else
\begin{figure}[t]
    \centering
    \includegraphics[width=\textwidth]{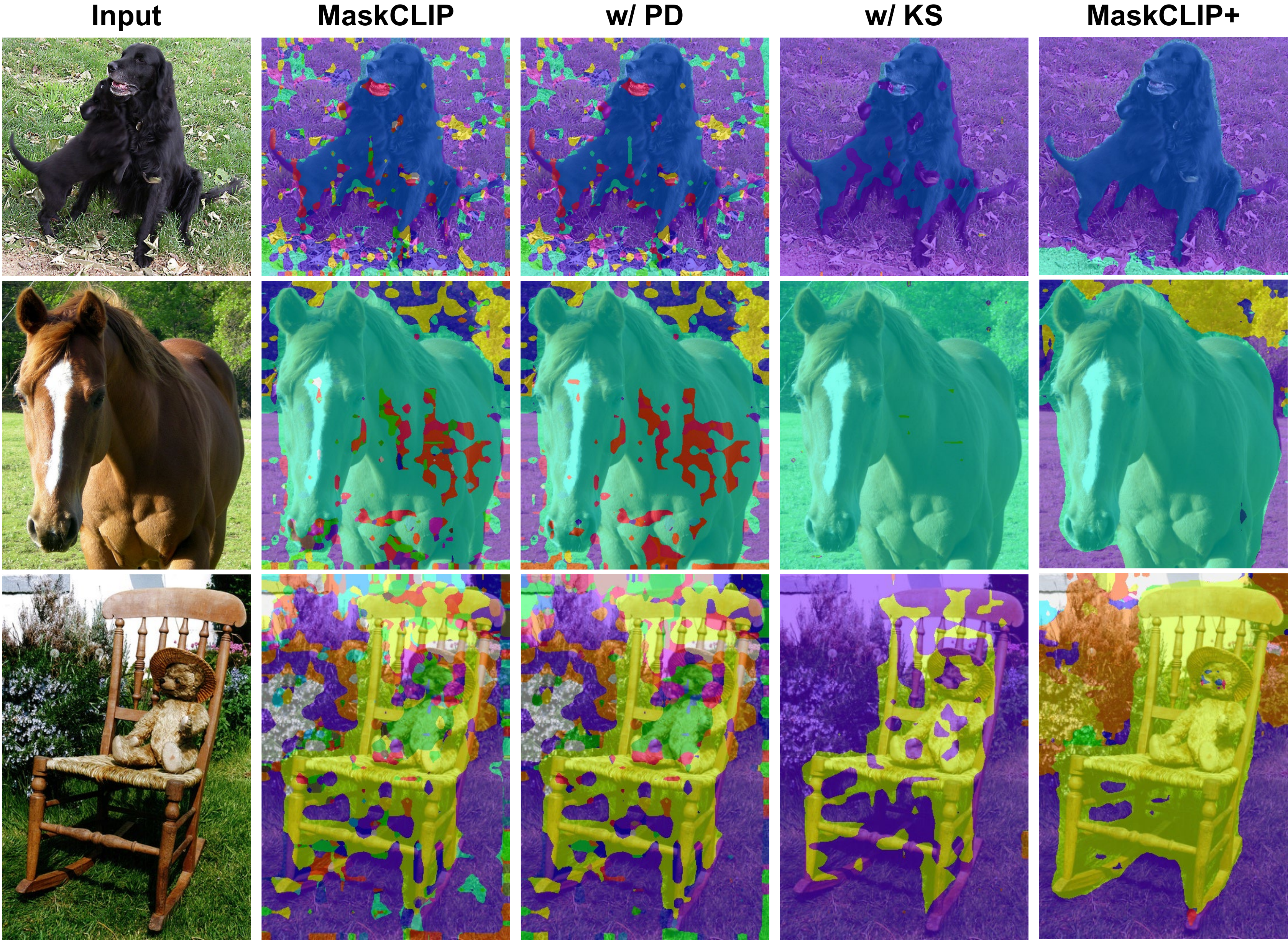}
    \caption{\inlinesection{More qualitative results on PASCAL Context.} Here all results are obtained \textbf{without} any annotation. PD and KS refer to prompt denoising and key smoothing respectively. Row 2, Col 4 shows a failure case of KS, where all the pixels in the image are labeled as the \emph{horse}. Note that, PASCAL Context does not contain \emph{bear} or \emph{teddy bear} classes and \methodname{} predicts the \emph{teddy bear} pixels as \emph{bedclothes}}
    \label{fig:supp-quality-abl}
    \vspace{-7pt}
\end{figure}

\fi

\section{Qualitative Results on Annotation-Free Segmentation}
In Figure~\ref{fig:supp-quality-abl}, we show more qualitative results of \methodname{} on the PASCAL Context dataset in the annotation-free setting. The results are consistent with our analysis in the main submission, where prompt denoising (PD) removes the unconfident distraction classes, key smoothing (KS) aggressively smooths the noisy predictions, and \methodname+ yields the best results through pseudo-label-training. We find the predictions of KS are often dominated by a few classes and we show a failure case in the Figure~\ref{fig:supp-quality-abl} (Row 2, Col 4), where one class dominates the whole image. Moreover, the behavior of \methodname{} and \methodname+ in Row 3 is interesting. Since the PASCAL Context dataset does not contain \emph{bear} or \emph{teddy bear} classes, \methodname{} classifies the \emph{teddy bear} pixels into \emph{bedclothes}, which is the most related class. Meanwhile, through pseudo-label-training, after observing the true \emph{bedclothes} pixels, \methodname+ decides to treat the teddy bear as part of the chair that it sits on.
% Finally, we find that incorporating KS and PD into MaskCLIP+ results in only marginal gains. We conjecture that, while the inductive bias introduced by KS and PD improves the lower bound of the performance, it also potentially limits the upper bound. As shown in Figure~\ref{fig:supp-quality-abl}, despite KS yields much less noisy output, it reduces the output diversity at the same time.

\begin{figure}[t]
    \centering
    \includegraphics[width=\textwidth]{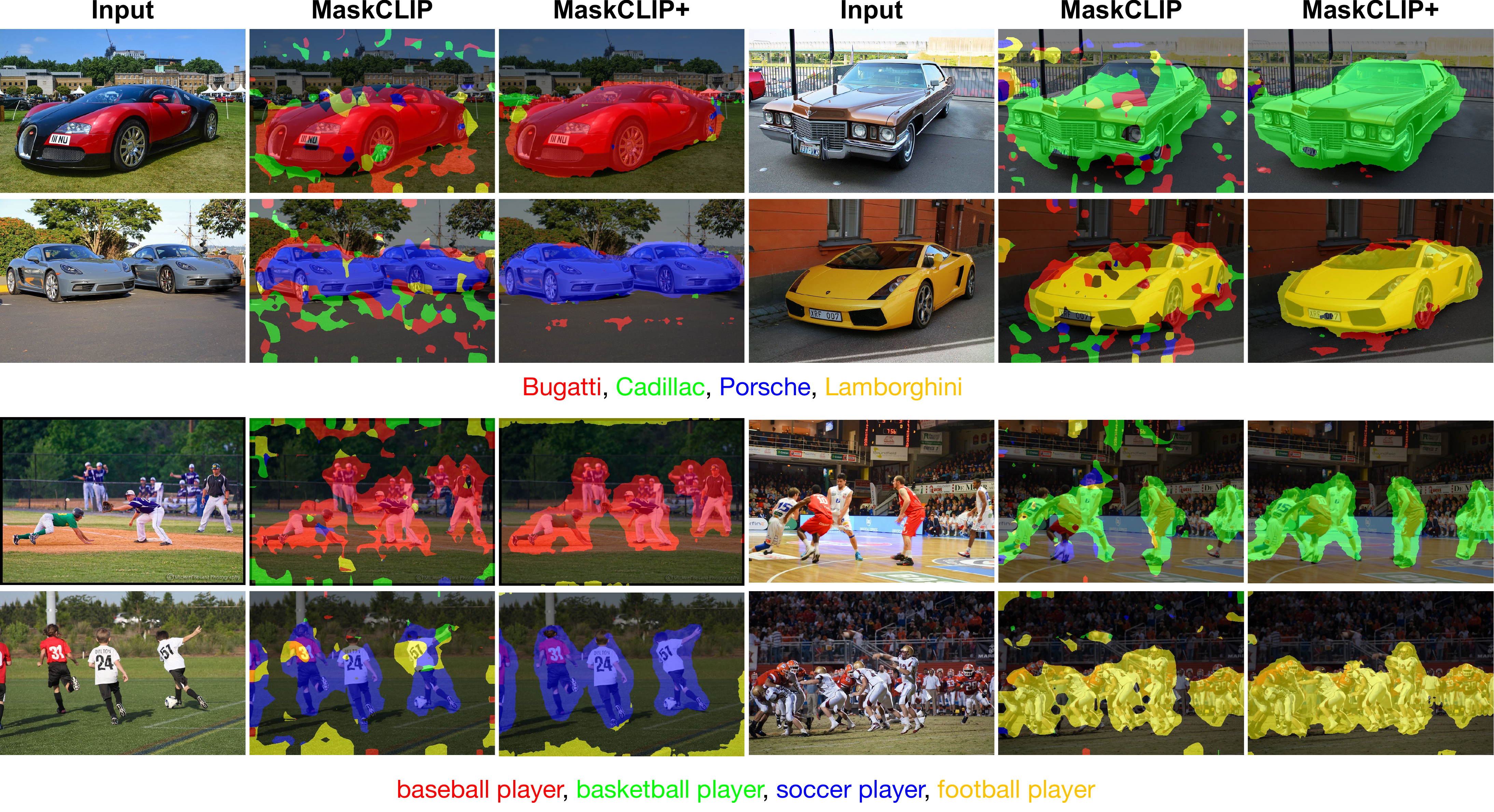}
    \caption{\inlinesection{More qualitative results on Web images.} \methodname~and \methodname+ can yield reasonable segmentation results of different car brands and sports \textbf{without} any annotation}
    \label{fig:supp-quality}
    \ifsupp
        \vspace{-9pt}
    \fi
\end{figure}

In our main submission, we show qualitative results of fine-grained classes (\emph{red car, yellow car}), objects with certain imagery properties (\emph{blurry car}), and novel concepts (\emph{Batman, Bill Gates}). Since \methodname~preserves the open-vocabulary ability, we can evaluate it on many interesting setups. In Figure~\ref{fig:supp-quality} we test whether \methodname~can segment out different car brands and sports. Similar to our main submission, the evaluation images are crawled from Flickr and all results are obtained without any annotation. \methodname~and \methodname+ again demonstrate powerful open-vocabulary ability on subtle concepts. Note that, in the \emph{basketball} and \emph{football} examples, \methodname~not only correctly distinguishes athletes playing different sports, but also separates audience and players.

\vspace{-10pt}

%%%%%%%% ROBUSTNESS RESULTS %%%%%%%%
% \newpage
\section{Robustness Results on Annotation-Free Segmentation}
In our main submission, we test the robustness of \methodname~under artificial corruptions. We use corrupting operations provided by the official code of ImageNet-C~\cite{imagenet-c}. In particular, the severity levels are controlled by a series of coefficients of corruption operators. Limited by space, in the main submission, we only include results of level 1 and level 5. Here, we extend the table to all levels. As shown in Table~\ref{tab:corrupt-supp}, CLIP-ViT-B/16 consistently outperforms CLIP-ResNet-50 by large margins and shows decent robustness.

We also supplement a baseline for the robustness test to compare with Table~\ref{tab:bad-input} in our main submission. In particular, we train an FCN segmentation model with the ViT-B/16 backbone (initialized with ImageNet-21K pre-trained weights) on PASCAL Context in a fully supervised manner for 40K iterations, then test the model on corrupted inputs. Table~\ref{tab:robust-supp} shows that MaskCLIP performs particularly well on Gaussian/shot/impulse noises.

\begin{table}[t]
\footnotesize
\centering
{
\caption{\inlinesection{More robustness results.} Here we evaluate \methodname~on PASCAL Context in the annotation-free setting under ImageNet-C corruptions across all severity levels. Results are reported in the mIoU metric}
\label{tab:corrupt-supp}
\begin{tabular}{l c cc c cc c cc c cc c cc}
    \noalign{\smallskip}
    \toprule
    \multirow{2}{*}{Corruption} && \multicolumn{2}{c}{\textbf{level 1}}  && \multicolumn{2}{c}{\textbf{level 2}}  && \multicolumn{2}{c}{\textbf{level 3}}  && \multicolumn{2}{c}{\textbf{level 4}}  && \multicolumn{2}{c}{\textbf{level 5}}  \\
                            && r50 & vit16      && r50 & vit16      && r50 & vit16      && r50 & vit16      && r50 & vit16 \\
    \midrule
    None                    && 18.5  & 21.7     && 18.5  & 21.7     && 18.5  & 21.7     && 18.5  & 21.7     && 18.5  & 21.7 \\
    \midrule
    Gaussian Noise          && 13.7  & 19.6     && 11.2  & 17.7     &&  7.9  & 14.8     &&  4.7  & 11.1     &&  2.1  & 6.8 \\
    Shot Noise              && 14.0  & 19.6     && 11.0  & 17.6     &&  7.8  & 14.8     &&  4.0  & 10.4     &&  2.4  & 7.5 \\
    Impulse Noise           &&  9.9  & 17.3     &&  8.1  & 15.9     &&  6.7  & 14.4     &&  4.1  & 10.9     &&  2.1  & 7.2 \\
    Speckle Noise           && 15.1  & 20.0     && 13.6  & 19.0     &&  9.6  & 16.0     &&  7.6  & 14.0     &&  5.6  & 11.4 \\
    \midrule
    Gaussian Blur           && 17.4  & 21.6     && 14.4  & 20.4     && 11.1  & 18.9     &&  8.1  & 17.3     &&  4.3  & 14.1 \\
    Defocus Blur            && 15.7  & 20.8     && 14.0  & 20.1     && 10.9  & 18.6     &&  8.5  & 17.1     &&  6.6  & 15.5 \\
    \midrule
    Spatter                 && 17.1  & 20.5     && 13.0  & 17.9     && 10.9  & 16.4     && 10.1  & 14.5     &&  7.8  & 12.2 \\
    JPEG                    && 15.7  & 20.8     && 14.3  & 20.1     && 13.3  & 19.5     && 10.3  & 17.4     &&  7.6  & 14.5 \\
    \bottomrule
\end{tabular}
}
\vspace{-7pt}
\end{table}
\begin{table}[t]
\footnotesize
\centering
{
\caption{\inlinesection{Baselines for Robustness Test.} Evaluation on PASCAL Context under level 5 corruptions with the ViT-B/16 backbone. N.: Noise, B.: Blur}
\label{tab:robust-supp}
\begin{tabular}{l c c c c c c c c c c}
    \noalign{\smallskip}
    \toprule
                & None            & Gauss N.           & Shot       & Impulse    & Speckle        & Gauss B.     & Defocus     & Spatter   & JPEG \\
    \midrule
    MaskCLIP    & 21.7            & \textbf{6.8}        & \textbf{7.5}  & \textbf{7.2}  & 11.4              & 14.1          & 15.5              & 12.2               & 14.5 \\
    Fully Sup.  & \textbf{54.5}   & 5.1                 & 6.7           & 4.8           & \textbf{22.7}     & \textbf{37.1} & \textbf{40.1}     & \textbf{31.5}      & \textbf{39.8} \\
    \bottomrule
\end{tabular}
}
\vspace{-5pt}
\end{table}

%%%%%%%% MORE EVALUATION METRICS %%%%%%%%
% \newpage
\section{Quantitative Results on Zero-Shot Segmentation}

In Table \ref{tab:supp-zero-shot-seen}, we report the mIoUs on seen classes of various methods. As mentioned in our main submission, across three standard datasets, using pseudo labels as the guidance, instead of distillation by feature matching, does not affect \methodname+'s performance on seen classes.

Apart from Intersection over Union (IoU), some zero-shot segmentation methods also report pixel accuracy (pAcc) and mean accuracy (mAcc) as evaluation metrics. For comprehensive comparisons, we provide performance with the mentioned metrics in Table \ref{tab:supp-zero-shot}. In terms of the overall and unseen pAcc/mAcc, \methodname+ still surpasses the previous SOTA methods by large margins and reaches near the fully-supervised baselines. However, its pAcc/mAcc of seen classes on PASCAL VOC and COCO-Stuff fall behind SPNet and CaGNet+ST by a bit. Different from mIoU, pAcc and mAcc punish only false negatives but not false positives (mIoU punishes both). Previous methods are much more confident on seen classes than unseen classes, therefore yields more predictions on seen classes, which consequently avoids false negatives on seen classes. In fact, SPNet biases towards seen classes so much that without calibration (reduce the confidence of seen classes by scaling factors), its performance on unseen classes is almost zero. \methodname+, on the contrary, is more balanced between seen and unseen classes.

\begin{table}[t]
\footnotesize
\centering
\def\sp{SPNet}
\def\spc{SPNet-C}
\def\zs{ZS3Net}
\def\cag{CaGNet}
\def\spst{SPNet}
\def\zsst{ZS3Net}
\def\cagst{CaGNet}
\def\strict{STRICT}
\def\denseclip{\methodname{}}
\def\denseclipplus{\methodname+}
\def\voc{\textbf{PASCAL-VOC}}
\def\context{\textbf{PASCAL-Context}}
\def\stuff{\textbf{COCO-Stuff}}
\newcommand{\better}[1]{\textcolor{ForestGreen}{(+#1)}}
\newcommand{\worse}[1]{\textcolor{BrickRed}{(-#1)}}

\caption{\inlinesection{Zero-shot segmentation performances on seen classes (mIoU)} 
% ST stands for self-training. mIoUs on seen classes are reported as the metric. \spc~is the SPNet with calibration. 
% The background class in PASCAL VOC is ignore, while that in PASCAL Context is not ignored. 
% On PASCAL Context, all methods use DeepLabv3+-ResNet101 as the backbone segmentation model and the rest two datasets use DeepLabv2-ResNet101. For \methodname+, CLIP-ResNet-50 is used to generate pseudo labels
}
\label{tab:supp-zero-shot-seen}
\begin{tabular}{l cc cc cc}
        \noalign{\medskip}
        \toprule
        Method                  && \voc      && \stuff       && \context \\
        % \midrule
        % Inductive \\
        \midrule
        \sp                     && 75.8      && 34.6         && \ph \\
        \spc                    && 78.0      && 35.2         && \ph \\
        \zs                     && 77.3      && 34.7         && 20.8 \\
        \cag                    && 78.4      && 35.5         && 24.8 \\
        % \midrule
        % Transductive \\
        \midrule
        \spst                   && 77.8      && 34.6         && \ph \\
        \zsst                   && 78.0      && 34.9         && 27.0 \\
        \cagst                  && 78.6      && 35.6         && \ph \\
        \strict                 && 82.7      && 35.3         && \ph \\
        \denseclipplus          && \textbf{88.8} 
                                && \textbf{38.2} 
                                && \textbf{44.4} \\
                                && \better{6.1}   
                                && \better{2.9}
                                && \better{17.4} \\
        \midrule
        Fully Sup.        && 88.6      && 38.1         && 44.4 \\
        \bottomrule
    \end{tabular}
\end{table}
\begin{table}[t]
\footnotesize
\centering
{
\def\sp{SPNet}
\def\spc{SPNet-C}
\def\zs{ZS3Net}
\def\cag{CaGNet}
\def\spst{SPNet}
\def\zsst{ZS3Net}
\def\cagst{CaGNet}
\def\strict{STRICT}
\def\denseclip{\methodname{}}
\def\denseclipplus{\methodname+}
\def\voc{\textbf{PASCAL-VOC}}
\def\context{\textbf{PASCAL-Context}}
\def\stuff{\textbf{COCO-Stuff}}
\newcommand{\better}[1]{\textcolor{ForestGreen}{(+#1)}}
\newcommand{\worse}[1]{\textcolor{BrickRed}{(-#1)}}

\caption{\inlinesection{Zero-shot segmentation performances (pAcc \& mAcc)}}
\label{tab:supp-zero-shot}
\setlength{\tabcolsep}{1pt}

\begin{tabular}{l ccc c ccc c ccc}
    \noalign{\medskip}
    \toprule
    \multirow{2}{*}{Method} & \multicolumn{3}{c}{\voc}     && \multicolumn{3}{c}{\stuff}       && \multicolumn{3}{c}{\context} \\
                            &  pAcc\textsubscript{(S)} & pAcc\textsubscript{(U)} & pAcc 
                            && pAcc\textsubscript{(S)} & pAcc\textsubscript{(U)} & pAcc 
                            && pAcc\textsubscript{(S)} & pAcc\textsubscript{(U)} & pAcc \\
    % \midrule
    % Inductive \\
    \midrule
    \sp                     & \textbf{94.8} & 0.0  & 76.9    && 65.6 & 1.7 & 51.3        && \ph & \ph & \ph \\
    \spc                    & 88.8 & 29.6 & 77.6    && 61.8 & 24.5 & 53.4       && \ph & \ph & \ph \\
    \zs                     & 93.0 & 21.5 & 79.4    && 64.3 & 22.8 & 54.7       && 53.5 & 58.6 & 52.8 \\
    \cag                    & 89.5 & 43.0 & 80.7    && 65.6 & 25.5 & 56.6       && 55.2 & 66.8 & 56.6 \\
    % \midrule
    % Transductive \\
    \midrule
    % \spst                   & \ph & \ph & \ph       && \ph & \ph & \ph          && \ph & \ph & \ph \\
    \zsst                   & 91.9 & 34.1 & 81.0    && 65.8 & 24.9 & 56.3       && 46.8 & 70.2 & 49.5 \\
    \cagst                  & 87.0 & 58.6 & 81.6    && \textbf{65.9} & 26.7 & 56.8       && \ph & \ph & \ph \\
    % \strict                 & \ph & \ph & \ph       && \ph & \ph & \ph          && \ph & \ph & \ph \\
    \denseclipplus          & 94.6 & \textbf{91.4} & \textbf{94.0}
                            && 64.2 & \textbf{79.4} & \textbf{67.6}
                            && \textbf{73.9} & \textbf{82.3} & \textbf{74.8} \\
                            & \worse{0.2}  & \better{32.8} & \better{12.4}    
                            && \worse{1.7}  & \better{52.7} & \better{10.8}        
                            && \better{18.7} & \better{12.1} & \better{18.2} \\
    \midrule
    Fully Sup.        & \ph & \ph & 94.0        && \ph & \ph & 68.1            && \ph & \ph & 74.8 \\
    \bottomrule
\end{tabular}

\vspace{10pt}

\begin{tabular}{l ccc c ccc c ccc}
    \toprule
    \multirow{2}{*}{Method} & \multicolumn{3}{c}{\voc}     && \multicolumn{3}{c}{\stuff}       && \multicolumn{3}{c}{\context} \\
                            &  mAcc\textsubscript{(S)} & mAcc\textsubscript{(U)} & mAcc 
                            && mAcc\textsubscript{(S)} & mAcc\textsubscript{(U)} & mAcc 
                            && mAcc\textsubscript{(S)} & mAcc\textsubscript{(U)} & mAcc \\
    % \midrule
    % Inductive \\
    \midrule
    \sp                     & \textbf{94.6} & 0.0  & 70.9    && 50.3 & 0.0  & 45.9         && \ph & \ph & \ph \\
    \spc                    & 87.9 & 23.9 & 71.9    && 46.3 & 16.1 & 43.6         && \ph & \ph & \ph \\
    \zs                     & 87.7 & 15.8 & 73.5    && 50.4 & 27.0 & 48.4         && 23.8 & 43.2 & 27.0 \\
    \cag                    & 88.7 & 39.4 & 76.4    && 50.7 & 27.0 & 48.5         && 35.7 & 49.8 & 36.8 \\
    % \midrule
    % Transductive \\
    \midrule
    % \spst                   & \ph & \ph & \ph       && \ph & \ph & \ph           && \ph & \ph & \ph \\
    \zsst                   & 85.7 & 26.4 & 73.8    && 50.4 & 27.2 & 48.6        && 32.3 & 57.1 & 36.4 \\
    \cagst                  & 83.9 & 50.7 & 75.6    && 50.6 & 27.3 & 48.5        && \ph & \ph & \ph \\
    % \strict                 & \ph & \ph & \ph       && \ph & \ph & \ph           && \ph & \ph & \ph \\
    \denseclipplus          & 93.7 & \textbf{92.6} & \textbf{93.4}
                            && \textbf{50.8} & \textbf{72.4} & \textbf{52.7}
                            && \textbf{55.4} & \textbf{80.0} & \textbf{59.5} \\
                            & \worse{0.9}  & \better{41.9} & \better{17.0}    
                            && \better{0.1}  & \better{45.1} & \better{4.1}        
                            && \better{19.7} & \better{22.9} & \better{22.7} \\
    \midrule
    Fully Sup.        & \ph & \ph & 93.4        && \ph & \ph & 53.0            && \ph & \ph & 59.5 \\
    \bottomrule
\end{tabular}

}
\end{table}
\clearpage

%%%%%%%% TARGET VOCABULARY %%%%%%%%
\begin{figure}[t]
\centering
\includegraphics[width=0.3\textwidth]{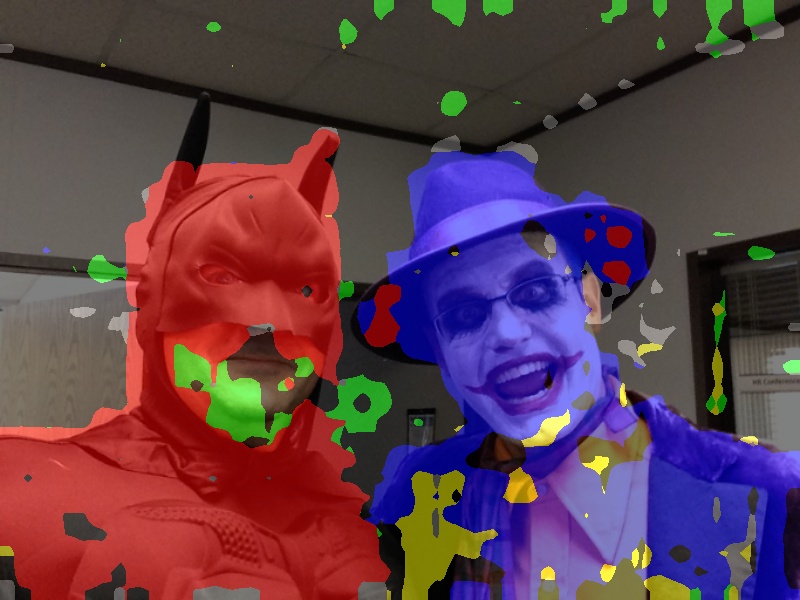}
\caption{\inlinesection{Open-vocabulary segmentation with a larger target text set}} 
\label{fig:more-text-supp}
\end{figure}
\section{Vocabulary Used in Open-Vocabulary Segmentation}
In Figure~\ref{fig:teaser} and Figure~\ref{fig:quality} of our main submission, all images share the same background classes, i.e., \textit{building, ground, grass, tree, sky}. For foreground classes, different images have a different set of targets, which are shown right below each image in Figure 4. In Figure~\ref{fig:more-text-supp}, we supplement an example with a larger vocabulary, with \textit{Batman, Joker, James Gordon, The Penguin, Robin, Alfred, Catwoman, Harley Quinn} as the foreground and all classes except \textit{person} in the Cityscapes as the background. We observe that Batman's jaw is segmented as \textit{James Gordon} and part of Joker's suit is classified into \textit{The Penguin}. Since certain local features are shared among multiple characters, it reveals that sometimes MaskCLIP cannot see broadly enough.

%%%%%%%% MULTI-SCALE INPUT ENSEMBLE %%%%%%%%
\begin{table}[t]
\footnotesize
\centering
{
\caption{\inlinesection{Input resolutions and multi-scale ensemble.} Here, we evaluate MaskCLIP on the PASCAL Context dataset}
\label{tab:multi-scale-supp}
\begin{tabular}{l cc cc cc cc cc}
    \noalign{\medskip}
    \toprule
    Input Res.  & 224   && 336   && 520       && [224, 520]    && [224, 336, 520] \\
    \midrule
    mIoU        & 22.72 && 23.02 && 21.68     && 25.16         && \textbf{26.34} \\
    \bottomrule
\end{tabular}
}
\end{table}
\section{Input Resolution and Multi-Scale Ensemble}
There is a trade-off in terms of the input resolution of MaskCLIP. Using the same input resolution as CLIP (224x224) assures the resolution/positional encoding matching but at the cost of yielding smaller output. We empirically find there exists a sweet spot at 336x336. We also find that multi-scale ensembles mitigate the resolution problem. (See Table~\ref{tab:multi-scale-supp}.)

%%%%%%%% ALGORITHM %%%%%%%%
\section{Pseudo Code of \methodname+}
The complete training process of \methodname+ is illustrated in Algorithm \ref{alg}.

\clearpage
\SetKwComment{Comment}{/* }{ */}
\begin{algorithm}[t]
\caption{\methodname+ pseudo code}\label{alg}
$P \gets \text{\methodname~model}$\;
$T \gets \text{text embeddings of target classes}$\;
$V_1 \gets \text{target model initialized w/ IN pre-trained}$\;
% $ \gets $\;
$V_1 \gets \text{load}\ T\ \text{to classifier weights of}\ V_1$\;
$\mathcal{D} \gets \text{images for training}$\;
$N_g \gets \text{\methodname-guided learning iterations}$\;
$N_s \gets \text{self-training iterations}$\;

\For{$i = 1,2,\dots,N_g$}{
  $\hat{y} \gets \text{model prediction}\ V_i(\mathcal{D}_i)$\;
  $y \gets \text{pseudo labels from \methodname}\ P(\mathcal{D}_i)$\;
  $\mathcal{L} \gets \text{cross entropy loss}\ \mathcal{L}_{CE}(\hat{y},y)$\;
  $V_{i+1} \gets \text{SGD model update}$\;
}
\For{$j = N_g+1,N_g+2,\dots,N_g+N_s$}{
  $\hat{y} \gets \text{model prediction}\ V_j(\mathcal{D}_j)$\;
  $y \gets \text{self-generated pseudo labels}\ V_j(\mathcal{D}_j)$\;
  $\mathcal{L} \gets \text{cross entropy loss}\ \mathcal{L}_{CE}(\hat{y},y)$\;
  $V_{j+1} \gets \text{SGD model update}$\;
}
\end{algorithm}

% \clearpage
\bibliographystyle{eccv/splncs04}
\bibliography{main}

\begin{thebibliography}{10}
\providecommand{\url}[1]{\texttt{#1}}
\providecommand{\urlprefix}{URL }
\providecommand{\doi}[1]{https://doi.org/#1}

\bibitem{zs3}
Bucher, M., Vu, T.H., Cord, M., P{\'e}rez, P.: Zero-shot semantic segmentation.
  In: NeurIPS (2019)

\bibitem{coco-stuff}
Caesar, H., Uijlings, J., Ferrari, V.: Coco-stuff: Thing and stuff classes in
  context. In: CVPR (2018)

\bibitem{swav}
Caron, M., Misra, I., Mairal, J., Goyal, P., Bojanowski, P., Joulin, A.:
  Unsupervised learning of visual features by contrasting cluster assignments.
  In: NeurIPS (2020)

\bibitem{naive-student}
Chen, L.C., Lopes, R.G., Cheng, B., Collins, M.D., Cubuk, E.D., Zoph, B., Adam,
  H., Shlens, J.: Naive-student: Leveraging semi-supervised learning in video
  sequences for urban scene segmentation. In: ECCV (2020)

\bibitem{deeplabv2}
Chen, L.C., Zhu, Y., Papandreou, G., Schroff, F., Adam, H.: Encoder-decoder
  with atrous separable convolution for semantic image segmentation. In: ECCV
  (2018)

\bibitem{deeplabv3+}
Chen, L.C., Zhu, Y., Papandreou, G., Schroff, F., Adam, H.: Encoder-decoder
  with atrous separable convolution for semantic image segmentation. In: ECCV
  (2018)

\bibitem{simclr}
Chen, T., Kornblith, S., Norouzi, M., Hinton, G.: A simple framework for
  contrastive learning of visual representations. In: ICML (2020)

\bibitem{cross-semi}
Chen, X., Yuan, Y., Zeng, G., Wang, J.: Semi-supervised semantic segmentation
  with cross pseudo supervision. In: CVPR (2021)

\bibitem{coco-caption}
Chen, X., Fang, H., Lin, T.Y., Vedantam, R., Gupta, S., Doll{\'a}r, P.,
  Zitnick, C.L.: Microsoft coco captions: Data collection and evaluation
  server. arXiv preprint  (2015)

\bibitem{simsiam}
Chen, X., He, K.: Exploring simple siamese representation learning. In: CVPR
  (2021)

\bibitem{imagenet}
Deng, J., Dong, W., Socher, R., Li, L.J., Li, K., Fei-Fei, L.: Imagenet: A
  large-scale hierarchical image database. In: CVPR (2009)

\bibitem{virtex}
Desai, K., Johnson, J.: Virtex: Learning visual representations from textual
  annotations. In: CVPR (2021)

\bibitem{pretask2}
Doersch, C., Gupta, A., Efros, A.A.: Unsupervised visual representation
  learning by context prediction. In: ICCV (2015)

\bibitem{pretask1}
Dosovitskiy, A., Springenberg, J.T., Riedmiller, M., Brox, T.: Discriminative
  unsupervised feature learning with convolutional neural networks. In: NeurIPS
  (2014)

\bibitem{pascal-voc}
Everingham, M., Eslami, S.A., Van~Gool, L., Williams, C.K., Winn, J.,
  Zisserman, A.: The pascal visual object classes challenge: A retrospective.
  IJCV  (2015)

\bibitem{self-cap}
Gomez, L., Patel, Y., Rusi{\~n}ol, M., Karatzas, D., Jawahar, C.:
  Self-supervised learning of visual features through embedding images into
  text topic spaces. In: CVPR (2017)

\bibitem{retrieval-cap}
Gordo, A., Larlus, D.: Beyond instance-level image retrieval: Leveraging
  captions to learn a global visual representation for semantic retrieval. In:
  CVPR (2017)

\bibitem{byol}
Grill, J.B., Strub, F., Altch{\'e}, F., Tallec, C., Richemond, P.H.,
  Buchatskaya, E., Doersch, C., Pires, B.A., Guo, Z.D., Azar, M.G., et~al.:
  Bootstrap your own latent: A new approach to self-supervised learning. In:
  NeurIPS (2020)

\bibitem{ViLD}
Gu, X., Lin, T.Y., Kuo, W., Cui, Y.: Open-vocabulary object detection via
  vision and language knowledge distillation. arXiv preprint  (2021)

\bibitem{context-seg}
Gu, Z., Zhou, S., Niu, L., Zhao, Z., Zhang, L.: Context-aware feature
  generation for zero-shot semantic segmentation. In: ACM MM (2020)

\bibitem{sbd}
Hariharan, B., Arbel{\'a}ez, P., Bourdev, L., Maji, S., Malik, J.: Semantic
  contours from inverse detectors. In: ICCV (2011)

\bibitem{moco}
He, K., Fan, H., Wu, Y., Xie, S., Girshick, R.: Momentum contrast for
  unsupervised visual representation learning. In: CVPR (2020)

\bibitem{resnet}
He, K., Zhang, X., Ren, S., Sun, J.: Deep residual learning for image
  recognition. In: CVPR (2016)

\bibitem{imagenet-c}
Hendrycks, D., Dietterich, T.: Benchmarking neural network robustness to common
  corruptions and perturbations. In: ICLR (2019)

\bibitem{clip-score}
Hessel, J., Holtzman, A., Forbes, M., Bras, R.L., Choi, Y.: Clipscore: A
  reference-free evaluation metric for image captioning. In: EMNLP (2021)

\bibitem{uncertain-seg}
Hu, P., Sclaroff, S., Saenko, K.: Uncertainty-aware learning for zero-shot
  semantic segmentation. In: NeurIPS (2020)

\bibitem{adv-semi}
Hung, W.C., Tsai, Y.H., Liou, Y.T., Lin, Y.Y., Yang, M.H.: Adversarial learning
  for semi-supervised semantic segmentation. In: BMVC (2018)

\bibitem{correct-semi}
Ibrahim, M.S., Vahdat, A., Ranjbar, M., Macready, W.G.: Semi-supervised
  semantic image segmentation with self-correcting networks. In: CVPR (2020)

\bibitem{label-prop}
Iscen, A., Tolias, G., Avrithis, Y., Chum, O.: Label propagation for deep
  semi-supervised learning. In: CVPR (2019)

\bibitem{diet-nerf}
Jain, A., Tancik, M., Abbeel, P.: Putting nerf on a diet: Semantically
  consistent few-shot view synthesis. In: ICCV (2021)

\bibitem{align}
Jia, C., Yang, Y., Xia, Y., Chen, Y.T., Parekh, Z., Pham, H., Le, Q.V., Sung,
  Y., Li, Z., Duerig, T.: Scaling up visual and vision-language representation
  learning with noisy text supervision. In: ICML (2021)

\bibitem{iccv-workshop}
Kato, N., Yamasaki, T., Aizawa, K.: Zero-shot semantic segmentation via
  variational mapping. In: ICCVW (2019)

\bibitem{pseudo-label}
Lee, D.H.: Pseudo-label: The simple and efficient semi-supervised learning
  method for deep neural networks. In: ICMLW (2013)

\bibitem{consistent-seg}
Li, P., Wei, Y., Yang, Y.: Consistent structural relation learning for
  zero-shot segmentation. In: NeurIPS (2020)

\bibitem{learn-self}
Li, X., Sun, Q., Liu, Y., Zhou, Q., Zheng, S., Chua, T.S., Schiele, B.:
  Learning to self-train for semi-supervised few-shot classification. In:
  NeurIPS (2019)

\bibitem{error-semi}
Mendel, R., De~Souza, L.A., Rauber, D., Papa, J.P., Palm, C.: Semi-supervised
  segmentation based on error-correcting supervision. In: ECCV (2020)

\bibitem{consist-semi}
Mittal, S., Tatarchenko, M., Brox, T.: Semi-supervised semantic segmentation
  with high-and low-level consistency. IEEE TPAMI  (2019)

\bibitem{pascal-context}
Mottaghi, R., Chen, X., Liu, X., Cho, N.G., Lee, S.W., Fidler, S., Urtasun, R.,
  Yuille, A.: The role of context for object detection and semantic
  segmentation in the wild. In: CVPR (2014)

\bibitem{pretask3}
Noroozi, M., Favaro, P.: Unsupervised learning of visual representations by
  solving jigsaw puzzles. In: ECCV (2016)

\bibitem{cross-consist-semi}
Ouali, Y., Hudelot, C., Tami, M.: Semi-supervised semantic segmentation with
  cross-consistency training. In: CVPR (2020)

\bibitem{cvpr-workshop}
Pastore, G., Cermelli, F., Xian, Y., Mancini, M., Akata, Z., Caputo, B.: A
  closer look at self-training for zero-label semantic segmentation. In: CVPRW
  (2021)

\bibitem{style-clip}
Patashnik, O., Wu, Z., Shechtman, E., Cohen-Or, D., Lischinski, D.: Styleclip:
  Text-driven manipulation of stylegan imagery. In: ICCV (2021)

\bibitem{meta-pseudo}
Pham, H., Dai, Z., Xie, Q., Le, Q.V.: Meta pseudo labels. In: CVPR (2021)

\bibitem{berkeley-cap}
Quattoni, A., Collins, M., Darrell, T.: Learning visual representations using
  images with captions. In: CVPR (2007)

\bibitem{clip}
Radford, A., Kim, J.W., Hallacy, C., Ramesh, A., Goh, G., Agarwal, S., Sastry,
  G., Askell, A., Mishkin, P., Clark, J., Krueger, G., Sutskever, I.: Learning
  transferable visual models from natural language supervision. In: ICML (2021)

\bibitem{denseclip}
Rao, Y., Zhao, W., Chen, G., Tang, Y., Zhu, Z., Huang, G., Zhou, J., Lu, J.:
  Denseclip: Language-guided dense prediction with context-aware prompting.
  arXiv preprint  (2021)

\bibitem{inverse}
Ravula, S., Smyrnis, G., Jordan, M., Dimakis, A.G.: Inverse problems leveraging
  pre-trained contrastive representations. In: NeurIPS (2021)

\bibitem{naver-cap}
Sariyildiz, M.B., Perez, J., Larlus, D.: Learning visual representations with
  caption annotations. In: ECCV (2020)

\bibitem{transformer}
Vaswani, A., Shazeer, N., Parmar, N., Uszkoreit, J., Jones, L., Gomez, A.N.,
  Kaiser, {\L}., Polosukhin, I.: Attention is all you need. In: NeurIPS (2017)

\bibitem{spnet}
Xian, Y., Choudhury, S., He, Y., Schiele, B., Akata, Z.: Semantic projection
  network for zero-and few-label semantic segmentation. In: CVPR (2019)

\bibitem{zsd-yolo}
Xie, J., Zheng, S.: Zsd-yolo: Zero-shot yolo detection using vision-language
  knowledge distillation. arXiv preprint  (2021)

\bibitem{transfer}
Yosinski, J., Clune, J., Bengio, Y., Lipson, H.: How transferable are features
  in deep neural networks? In: NeurIPS (2014)

\bibitem{contrast-cap}
Yuan, X., Lin, Z., Kuen, J., Zhang, J., Wang, Y., Maire, M., Kale, A., Faieta,
  B.: Multimodal contrastive training for visual representation learning. In:
  CVPR (2021)

\bibitem{open-parse}
Zhao, H., Puig, X., Zhou, B., Fidler, S., Torralba, A.: Open vocabulary scene
  parsing. In: ICCV (2017)

\bibitem{pspnet}
Zhao, H., Shi, J., Qi, X., Wang, X., Jia, J.: Pyramid scene parsing network.
  In: CVPR (2017)

\bibitem{pseudo-seg}
Zou, Y., Zhang, Z., Zhang, H., Li, C.L., Bian, X., Huang, J.B., Pfister, T.:
  Pseudoseg: Designing pseudo labels for semantic segmentation. In: ICLR (2021)

\end{thebibliography}
\end{document}